# A Weighted Mutual k-Nearest Neighbour for Classification Mining

Joydip Dhar, Ashaya Shukla, Mukul Kumar, Prashant Gupta
ABV-Indian Institute of Information Technology & Management
Gwalior-474015, M.P, India

*Abstract--* **kNN is a very effective Instance based learning method, and it is easy to implement. Due to heterogeneous nature of data, noises from different possible sources are also widespread in nature especially in case of large-scale databases. For noise elimination and effect of pseudo neighbours, in this paper, we propose a new learning algorithm which performs the task of anomaly detection and removal of pseudo neighbours from the dataset so as to provide comparative better results. This algorithm also tries to minimize effect of those neighbours which are distant. A concept of certainty measure is also introduced for experimental results. The advantage of using concept of mutual neighbours and distance-weighted voting is that, dataset will be refined after removal of anomaly and weightage concept compels to take into account more consideration of those neighbours, which are closer. Consequently, finally the performance of proposed algorithm is calculated.**

## I. INTRODUCTION

Data mining is the processing of data for different perspectives, in order to find out interesting patterns in large databases and compiling them into useful information. Recent years have attracted a significant amount of research in almost all the aspects of data mining, where pattern classification is one of the most fundamental and widely studied research topics in the field of data mining and classification. In the field of classification, a successful application of an algorithm relies very much on the quality of data. In practical situations, collection of data is often done from miscellaneous diverse data sources.

As a result of the ignorance of quality of data, erroneous decisions may be predicted by the classification learning algorithms. Moreover, there would be increased complexity, involved in the construction of the classification models. Hence, in order to implement classification algorithms correctly, development of an effective technique for removal of noise from the dataset has become essential. It is worthy to mention that noisy instances bring less impact to simple learning algorithms, such as $NBC$ and $kNN$ than sophisticated classifiers such as $SVM$ or random forests [4]. Due to the simplicity and easiness in implementation of $kNN$, this off-the-shelf method has also been applied to get rid of the noisy instances in databases, and can achieve competitive results as compared to the most sophisticated learning algorithms [1], [2], [5], [6].

In this work a non-parametric lazy learning algorithm $kNN$ ($k$-Nearest Neighbours) and its five more variants with variations in algorithmic procedure, where the performance of each algorithm depends upon the various factors is discussed. We propose a new learning algorithm which is going to perform the task of anomaly detection and providing weightage to relatively close neighbours in comparison with distant neighbours. Rest of the paper is organized as follows section-two comprises problem definition, section three methodology, section four results and discussion followed by future scope and conclusion in section five and six respectively.

## II. PROBLEM DEFINITION

Pattern classification provides intuition about any unseen data that what kind of behavior this data is going to show on the basis of patterns seen in history i.e., on the basis of historical data [2]. For a person it is generally easy to predict that which sound is the sound of a male and which one is of a female, differentiate between handwritten letters but when this comes to computer programs it is not the same situation. For a programmable computer it is difficult to solve these kinds of perceptual problems [9]. For this case, it is difficult because as each pattern comprises huge amount of information and recognition problems typically have a subtle, high dimensional structure [9]. Formally it can be argued that pattern classification is organization of patterns into groups of patterns sharing same kind of properties. It can be applied to many possible potentials but classification is most widely used and has attracted more attention of researchers.

Classification is basically a procedure in which one is intended towards defining a model or precisely it can be said that hypothesis function that learns from given dataset about behavior of data items on the basis of various features intended for able to predict the class label for upcoming unknown instances. Let $x_i$ be an input instance represented as p-dimensional vector form, i.e., $x_i = (x_{i1}, \ldots, x_{ip})$, $C = \{c_1, \ldots, c_m\}$ be a set of class labels. The class label of instance $x_i$ fits to one of the categories $c_i$, i.e., there is a scalar function $f$, which allocates a class label, $c_i = f(x_i)$, to every instance. Given a dataset $D$ consisting of $n$ pairs of instance and label, i.e., $D = \{(x_1, c_1), \ldots, (x_n, c_n)\}$. The classification task is to determine the scalar function (i.e., hypothesis) $f$, on which the class labels of unclassified instances can be predicted specifically [1]. Starting from most basic and influential work of Cover and Hart in 1967 [8], their proposed **1NN** was perhaps the most influential step towards classification and straightforward learning algorithm. The fundamental idea behind this algorithm was to take in consideration only one nearest neighbour for prediction of class label for any upcoming unknown instances. Formally let's say $x_i$ is an unknown instance a given a dataset $D$. In $1NN$ algorithm, the first job that is performed is to store all the instances of training set in memory, so that they can be used for further querying. For the prediction of the class label for the unknown instance $x_i$, searching of its nearest neighbour $x_0$ is performed, it will predict the class label of unknown instance $x_i$ as $c$ which is derived from $f(x_0)$. So class of the input instance $x_i$ will be the same class that is of $x_0$. $1NN$ only considers the information regarding the closest nearest neighbour, so it is more susceptible of noise and also decision boundary is very sharp. So there are various variants which try to minimize these limitations. In this work, we proposed a new algorithm that is intended towards performing well with almost all kinds of datasets.

### III. METHODOLOGY

As discussed in last section the most basic method used for classification is $1NN$. Irrespective of its simplicity and easiness in implementation, it has a major drawback of its susceptible behaviour towards noise and over fitting of decision boundary. A number of extensions of this algorithm came into existence. First and the most general kind of extension is $kNN$ i.e., extending one nearest neighbour to $k$ nearest neighbours. In case of $1NN$, only one nearest neighbour was taken into consideration, but in case of $kNN$, $k$ nearest neighbours are taken into consideration. Let $N_k(x)$ be the set of $k$ nearest neighbour of instance $x$. Prediction of class label of $x$ will be based on only majority voting among all the k nearest neighbours of instance $x$.

$Majority\ voting: y' = argmax_v \sum_{(x_i,y_i) \in D_z} I(v = y_i).$ (1)

In equation (1), $v$ is class label, $y_i$ is class label for one of the nearest neighbours $I(.)$ is an indicator function that returns value one when argument is correct otherwise it returns zero [3].

Despite of its simplicity, $kNN$ gives comparative results in comparison with other sophisticated algorithms of machine learning. Impact of each neighbour in case of $kNN$ is same, weights can be applied to each neighbours for classification, this extension of $kNN$ is named as weighted $kNN$ [9]. This Weighted $kNN$ (i.e., $WkNN$) makes use of Distance-Weighed voting in consideration for class label prediction.

$y' = argmax_v \sum_{(x_i,y_i) \in D_z} w_i \times I(v = y_i),$ (2)

where, $w_i$ is weights assigned to each neighbour.

In [1] Huawen Liu and Shichao Zhang (2011) argued that concept of mutual neighbours can be applied to remove anomalies from datasets i.e.,

$M_k(x) = \{x_i \in D | x_i \in N_k(x) \land x \in N_k(x_i)\}.$ (3)

This definition says that given an instance $x$, if $x_i$ comes in nearest neighbours of $x$ then $x$ should also be in the nearest neighbours of $x_i$. This concept led to the origin of Mutual $k$ Nearest Neighbour (i.e., $MkNN$) algorithm [1].

In this paper, a new algorithm is proposed by integrating weighted concept and concept of mutual neighbour together. This tries to remove anomaly and provides weightage to relatively close neighbour in case of class label prediction. This is termed as Weighted Mutual $kNN$ (i.e., $WMkNN$). The whole procedure of implementation of the proposed algorithm is as follows:

Steps for noise removal in training set:

1. For each instance $x$ in training set of size of $m$, compute distance of each instance with all other instances.

2. Sort distances for each instance $x$.

3. Obtain $k$ mutual nearest neighbours $M_k(x)$.

4. Remove all instances x with $M_k(x) = \emptyset$.

5. Reduced dataset obtained will be further used as training set.

Steps for class prediction of test example:

1. For an instance $x_t$, obtain $M_k(x_t)$.

2. If $M_k(x_t) \neq \emptyset$ class label of $x_t$ will be predicted using weightage of class labels of all instances in set $M_k(x_t)$.

3. If $M_k(x_t) = \emptyset$, $xt$ will be considered as an outlier and its class label will not be predicted. This step tries to provide more certain results in the prediction of class label.

The Flow diagram of the proposed algorithm is shown in Figure 1.

### IV. IMPLEMENTATION

Classification performance of a learning algorithm is the criterion on which the validity of the algorithm is judged. To validate our learning algorithm we took three standard dataset with different types and sizes. All the three datasets are freely accessible at the UCI Machine Learning Repository [7].

TABLE I. DATASETS USED.

| Datasets | Instances | features | classes |
|---|---|---|---|
| Glass | 214 | 10 | 7 |
| Wine | 178 | 14 | 3 |
| ILPD | 583 | 10 | 2 |

These three benchmark datasets are frequently used for validating classification purpose. In table 1, first column describes the name of the dataset and second describes number of instances in that dataset and third column represents how many features are there in those databases and last column represents number of classes in that particular database. These benchmark datasets are not biased, providing good importance in understanding algorithms and helps in comprehensive test in validation of learning algorithms. As our proposed method is an extension of $MkNN$ and $kNN*$ further $MkNN$ is an extension of $kNN$. So we implemented all these algorithms to make ourselves capable of making comparison among these algorithms. Moreover, comprehensive test and validation of these algorithms can only be made by comparing with already existing algorithms.

Apart from accuracy of different algorithms on benchmark datasets, we also tried to calculate certainty measure in case of every dataset with respect to all the six algorithms. Six algorithms are: $kNN$, $WkNN$, $MkNN$, $kNN*$, $WMkNN$, $WkNN*$. Results obtained are discussed in section 4.

V. RESULTS AND DISCUSSIONS

As mentioned in preceding section, the proposed algorithm is an extension of $kNN$ and $MkNN$ by integrating weightage concept in it. Thus, we took $kNN$, $WkNN$ and $MkNN$ as baseline and made a comparison among kNN, WkNN, MkNN and WMkNN. In order to show effectiveness of the proposed method used for noise removal kNN is also applied to the reduced training set named as $kNN*$. $KNN*$ has been also observed by integrating weightage concept and coined it as $WkNN*$.

We also used concept of certainty measure is to describe how much certain are the results. It can be derived as:

$$CM_i = \frac{Total\ votes\ (i)}{\sum_{c=1}^{\#of\ classes}(total\ votes(c))}. \quad (4)$$

In equation (4), $CM_i$ is certainty factor of any prediction '$i$' and total votes $(i)$ is frequency of the class $i$ in $k$ nearest neighbours. Weighted concept is integrated in certainty measure named Weighted-Certainty measure:

$$WCM_i = \frac{Total\ weight\ (i)}{\sum_{c=1}^{\#of\ classes} Total\ weight\ of\ class(c)}, \quad (5)$$

where, $WCM_i$ is weighted certainty measure of any prediction '$i$' and $total\ weight\ (i)$ is sum of all weights the class $i$ in $k$ nearest neighbours. Weight used is derived as :

$$weight = \frac{1}{d(x,y)}, \quad (6)$$

where, $d(x,y)$ is the distance between two instances $x$ and $y$.

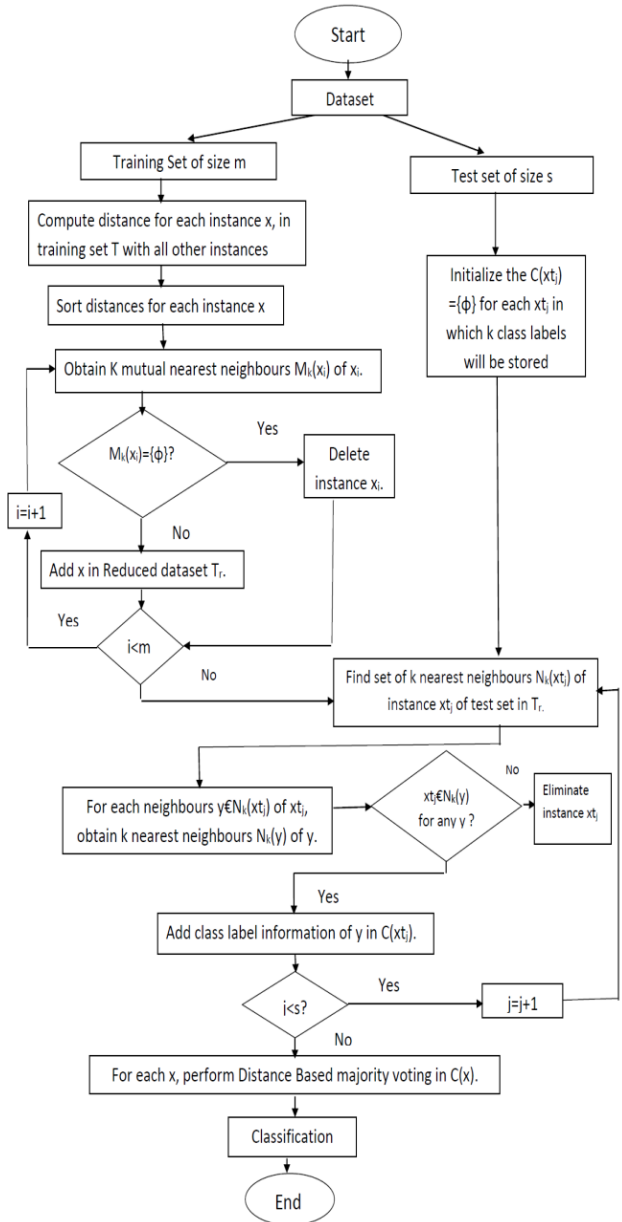

Fig. 1. Flow diagram of $WMkNN$.

TABLE II. ACCURACY VALUES FOR ILPD DATASET.

|  | $k=3$ | $k=4$ | $k=5$ | $k=6$ | $k=7$ |
| --- | --- | --- | --- | --- | --- |
| $kNN$ | 60.09 | 61.66 | 61.30 | 62.47 | **63.24** |
| $WkNN$ | 62.86 | 60.86 | 62.09 | 61.69 | **64.07** |
| $kNN*$ | 58.92 | 60.46 | 60.90 | 62.47 | **63.24** |
| $WkNN*$ | 61.29 | 60.44 | 62.09 | 61.69 | **64.07** |
| $MkNN$ | 58.16 | 60.05 | 62.23 | 63.54 | **65.34** |
| $WMkNN$ | 58.63 | 59.16 | 62.21 | 63.59 | **65.37** |

The experimental results of all the databases are compared on three basis:
1. Comparison among $kNN$, $kNN*$, $MkNN$.
2. Comparison between weighted and Non-Weighted algorithms.
3. Comparison for different values of $k$.

For the results in table 2, table 3, table 4 and table 5, value of $k$ is assigned in between three and seven. It is notable from results observed in table 2 and table 4 that $WkNN$ is performing better than $kNN$ is almost all the cases. For example, in case of ILPD dataset, for $k = 3$ $WkNN$, $WkNN*$ and $WMkNN*$ is performing better than $kNN$, $KNN*$ and $WMkNN$ respectively.

The experimental results on ILPD dataset are given in table 2. It can be noticed that the accuracy of $MkNN$ is not always better than $kNN$ and $kNN*$. For example, for $k$=3 and $k$=4 accuracy of $kNN$ and $kNN*$ is better than $MkNN$ for the given dataset. It has been observed that the accuracy of $MkNN$ is greater than $kNN$ and $kNN*$ for all other values of $k$, Since some of the instances of test data is eliminated while predicting the class labels of test data set using $MkNN$. But the accuracy of $kNN*$ is lesser than $kNN$ in almost all the cases. Observations shows that for different $k$ values, algorithms are performing differently. For example, in table 2, at $k$=3 $kNN$ is providing best results, secondly $kNN*$ and lastly $MkNN$. But overall if average accuracies of all the algorithms for different values of $k$ is computed, then it is noticeable that for $k$=7, they are coming up with best results. It is observed from the result table that certainty measure directly depends on the value of k. With increment in value of $k$, certainty decreases. The results of certainty measure on GLASS dataset is described in table 5.

TABLE III. CERTAINTY VALUES FOR ILPD DATASET.

|  | $k = 3$ | $k = 4$ | $k = 5$ | $k = 6$ | $k = 7$ |
|---|---|---|---|---|---|
| $kNN$ | **79.39** | 76.74 | 73.95 | 73.55 | 72.56 |
| $WkNN$ | **80.38** | 76.75 | 74.49 | 73.63 | 72.86 |
| $kNN*$ | **78.52** | 75.76 | 73.47 | 73.31 | 72.57 |
| $WkNN*$ | **79.64** | 75.83 | 74.12 | 73.38 | 72.87 |
| $MkNN$ | **85.92** | 81.30 | 78.39 | 77.92 | 77.40 |
| $WMkNN$ | **85.92** | 81.32 | 78.42 | 77.93 | 77.45 |

TABLE IV. ACCURACY VALUES FOR GLASS DATASET

|  | $k = 3$ | $k = 4$ | $k = 5$ | $k = 6$ | $k = 7$ |
|---|---|---|---|---|---|
| $kNN$ | 69.13 | 68.18 | **70.08** | 67.77 | 68.70 |
| $WkNN$ | **70.08** | 68.38 | 70.02 | 68.65 | 69.13 |
| $kNN*$ | 66.83 | 67.72 | **70.54** | 67.77 | 68.24 |
| $WkNN*$ | 67.79 | 67.72 | **69.56** | 67.74 | 68.22 |
| $MkNN$ | 72.32 | **74.31** | 72.68 | 72.38 | 71.59 |
| $WMkNN$ | 73.47 | **74.19** | 72.07 | 70.77 | 70.53 |

In case of Certainty measure, in table 3, it is notable that $MkNN$ is providing better results than $kNN$ and $kNN*$ in all the cases, because in $MkNN$, prediction of the class label of those instances, which are suspicious are not being done. While on an average $kNN$ is providing more certain results than $kNN*$. From table 3 it is also notable that the weighted algorithms are providing more certain results than the corresponding non-weighted algorithms. For example, for $k$=3, $WMkNN$ performs exceptionally well certain results than other algorithms.

Our proposed algorithm $WMkNN$ is giving creditable results for almost all the datasets. This proposed algorithm is determining class labels using concept of mutual neighbour, that is used for removal of noise and also assigning weights so that relatively close neighbours will be taken into more consideration. It is notable from results that $MkNN$ is providing better results. From the certainty table it was observed that certainty factor is associated with value of $k$. Increment in value of $k$ cause decrement in certainty. It was also noticeable that $MkNN$ and $WMkNN$ are giving more certain results, as in both the cases doubtful instances are deleted.

TABLE V. CERTAINTY VALUES FOR GLASS DATASET.

|  | $k = 3$ | $k = 4$ | $k = 5$ | $k = 6$ | $k = 7$ |
|---|---|---|---|---|---|
| $kNN$ | **84.67** | 80.32 | 76.61 | 75.55 | 74.23 |
| $WkNN$ | **84.88** | 80.35 | 77.58 | 76.31 | 74.78 |
| $kNN*$ | **84.14** | 79.18 | 77.03 | 76.19 | 74.18 |
| $WkNN*$ | **84.36** | 79.22 | 77.68 | 76.22 | 74.39 |
| $MkNN$ | **89.02** | 85.44 | 83.62 | 81.73 | 80.12 |
| $WMkNN$ | **89.28** | 85.79 | 83.72 | 81.84 | 80.27 |

VI. FUTURE SCOPE

Proposed learning algorithm which is termed as $WMkNN$ tries to minimize the effect of pseudo neighbours on prediction of class labels and provide weightage to closer neighbours in the set of k nearest neighbour. The distance metric we have used in our approach is Euclidian distance, proposed algorithm can be further improved by applying certain distance metric, which works well with datasets comprising mixture of numerical and nominal values and datasets with missing values. In case of large datasets, especially datasets having large number of features, correlation analysis can be done apart from concept of mutual neighbours. It may be possible to use variable values of k for skewed datasets for further enhancement in proposed algorithm.

VII. CONCLUSION

In this work, we proposed a new learning algorithms that does job of noise removal and apply weighted concept at the same time called $WMkNN$. The motivation behind this algorithm was to eliminate noisy instances and provide weightage to closer neighbours in the job of prediction of class labels for unknown instances. Precisely it removed noisy instances by the virtue of concept of mutual neighbour and provide more weightage to closer neighbour by virtue of distance-based voting. Besides, a variation of proposed algorithm was also been performed that used notion of mutual neighbour only for training set. An extensive analysis of some benchmark datasets of UCI machine learning database was done. Results of analysis showed that the new proposed algorithm were came up with better results than conventional $kNN$. It is observed from the experimental results that $WMkNN$ is performing better than $MkNN$ and $kNN$ for value of $k = 3$ in almost all the datasets, we used for our experimentation procedure.

Proposed certainty measure came up with conclusion that it is associated with the value of $k$ and as $k$ increases certainty decreases. It also concluded that certainty measure for $MkNN$ is better than $kNN$. The accuracy results provided shows that classification performance by $WMkNN$ is better than conventional $kNN$.